\documentclass[letterpaper,10pt,conference]{ieeeconf}
\usepackage{cite}
\usepackage{amsmath,amssymb,amsfonts}
\usepackage{algorithmic}
\usepackage{graphicx}
\usepackage{textcomp}
\usepackage{xcolor}
\usepackage{booktabs}
\usepackage{tabularx}
\usepackage{multirow}
\usepackage{xspace}
\usepackage{colortbl}
\usepackage{flushend}
\usepackage{svg}
\usepackage{balance}
\usepackage{comment}

\makeatletter
\g@addto@macro\@maketitle{%
  \begin{center}
    \vspace{-0.5em}
    \includegraphics[width=0.98\linewidth]{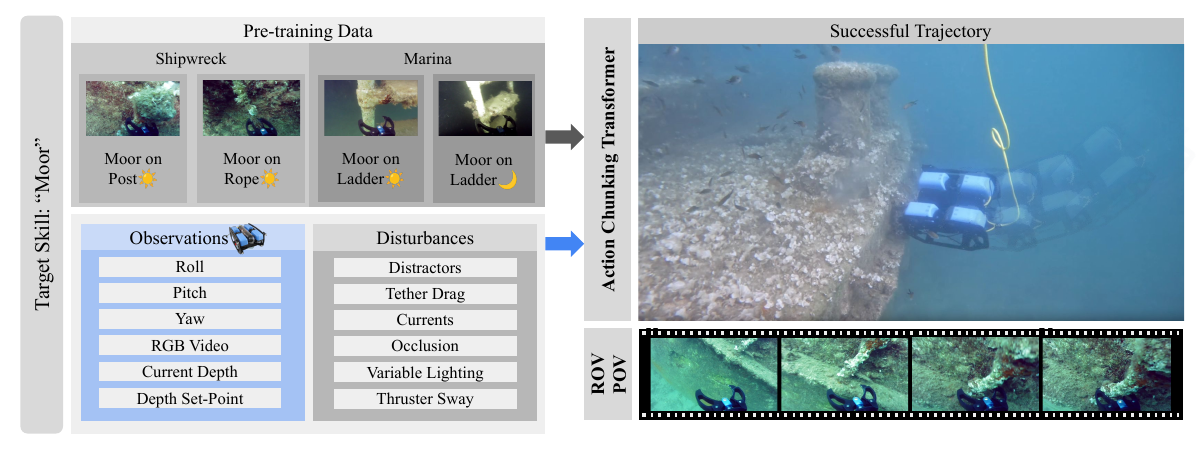}\\[0.6em]
    \setcounter{figure}{0}\refstepcounter{figure}%
    \parbox{\linewidth}{%
      \small\textbf{Fig.~\thefigure.} \textbf{Overview of Autonomous Underwater Mooring using Action Chunking Transformers (ACT).} 
      \textbf{(Left)} Training data collected across shipwreck and marina environments incorporating multi-modal observations and real-world disturbances. These include marine life distractors, tether drag, ocean currents, visual occlusion from water particulate, variable lighting based on depth or time of day, and thruster sway inherent to fluid dynamics. 
      \textbf{(Right)} Autonomous policy execution showing closed-loop mooring trajectory on the \textit{Coralita X127} shipwreck.
    }\label{fig:method}%
    \vspace{-2em}
  \end{center}
}
\makeatother

\begin{document}

\title{\LARGE\bfseries
Embodied at Sea: Imitation Learning for Multi-Task Underwater Manipulation}

\author{%
    Greta Perez-Haiek\textsuperscript{*,1}, Luca Macesanu\textsuperscript{*,1},
    Kevin Angers\textsuperscript{1}, Mohammed Imam\textsuperscript{1},\\
    Timothy Gambin\textsuperscript{\dag,2}, Christopher
  Clark\textsuperscript{\dag,1}\\[0.4em]
    {\small \textsuperscript{*}Equal contribution \quad
  \textsuperscript{\dag}Advising role}\\
    {\small \textsuperscript{1}New York University \quad
      \texttt{\{gp2583, lim2045, ka3658, mki4908, c.clark\}@nyu.edu}}\\
    {\small \textsuperscript{2}L-Universit\`a ta' Malta \quad
      \texttt{timmy.gambin@um.edu.mt}}\\
    {\small \texttt{https://lair-research-nyu.github.io/Embodied-at-Sea/}}%
  }

\maketitle

\newcommand{\ours}{\textsc{BluEmbodied Kit }}
\begin{abstract}
 Behavior Cloning has demonstrated promising generalization capabilities for robotic manipulation in terrestrial environments. However, marine visual distortions and dynamic currents create a severe domain shift that remains a key bottleneck in adapting generalist architectures to underwater field manipulation. To address this, 1616 episodes of underwater manipulation data were collected in calm and turbulent waters and were used to train 14 generalist Vision-Language-Action (VLA) models (DynamicVLA) with varying proprioception observation configurations. Across three tasks on 1260 gradable on-hardware rollouts in a pool, 10 models outperformed the baseline, with the highest-performing model achieving a 93.3\% overall success rate, including partial successes. Results are also introduced from a case study evaluating the performance of a single-task Action Chunking Transformer (ACT) and a generalist VLA (SmolVLA) policy for two tasks in marine archaeology and conservation: grasping and mooring. Experiments are performed on an ROV on and around the WWII wreck \textit{Coralita X127} and at Manoel Island Marina, Malta. Across 748 collected episodes and 508 gradable on-hardware rollouts, the ACT policy substantially outperformed SmolVLA (47.9\% vs.\ 7.9\% pooled success rate), earning a 66.7\% success rate on the most successful task at the marine archaeological site. To enable further contributions to embodied intelligence in marine field robotics, the \ours used to collect data, train models, and deploy policies on the ROV has been open sourced.

\end{abstract}

\begin{figure*}[t!]
  \centering
  \resizebox{0.95\textwidth}{!}{%
  \begin{tabular}{c c c c c c}

  & \textbf{Grab Green Rod}
  & \textbf{Grab Red Rod}
  & \textbf{Grab White Rod}
  & \textbf{Moor on Post}
  & \textbf{Remove Ring from Post} \\

  & \small\textit{(Pool)}
  & \small\textit{(Pool)}
  & \small\textit{(Pool)}
  & \small\textit{(Pool)}
  & \small\textit{(Pool)} \\[0.3em]

  $\vcenter{\hbox{\rotatebox[origin=c]{90}{\textbf{Initial State}}}}$ &
  $\vcenter{\hbox{\includegraphics[width=0.19\textwidth]{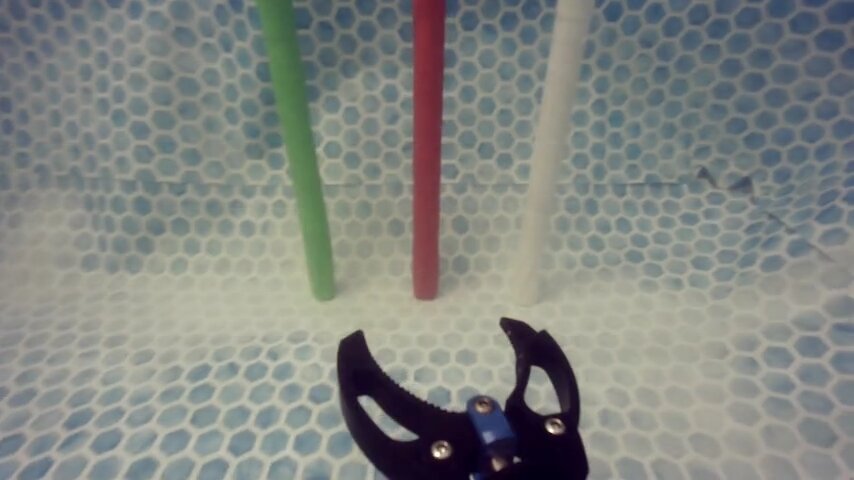}}}$ &
  $\vcenter{\hbox{\includegraphics[width=0.19\textwidth]{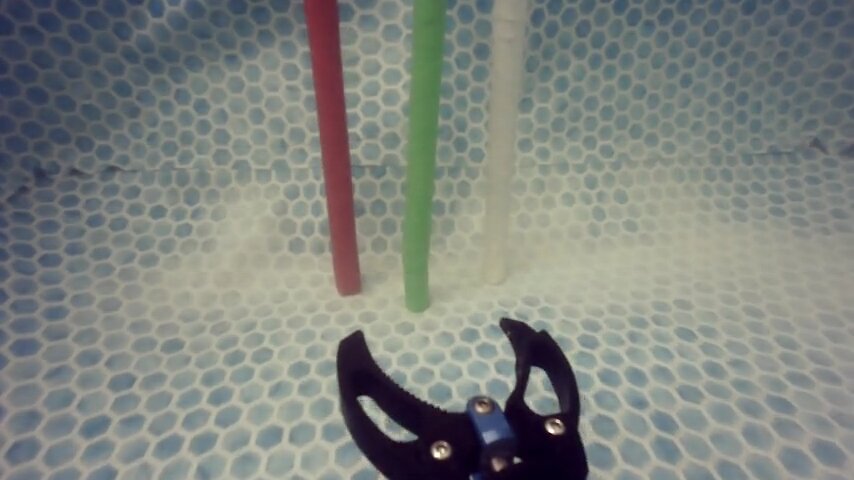}}}$ &
  $\vcenter{\hbox{\includegraphics[width=0.19\textwidth]{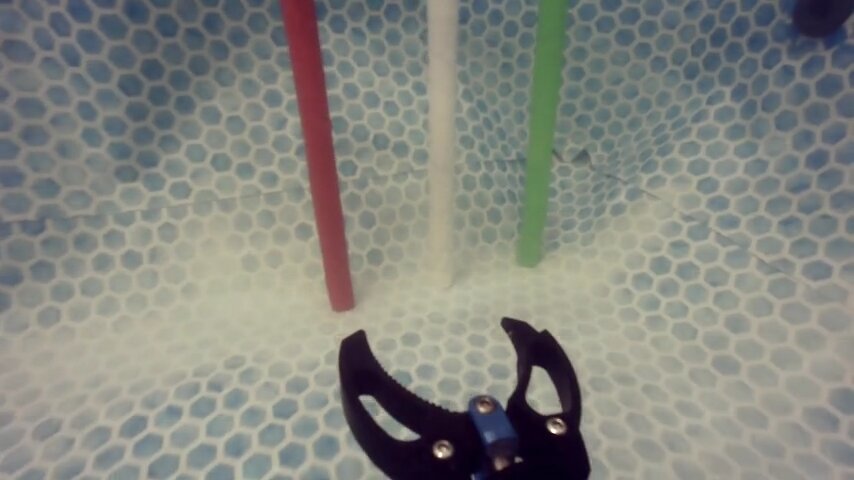}}}$ &
  $\vcenter{\hbox{\includegraphics[width=0.19\textwidth]{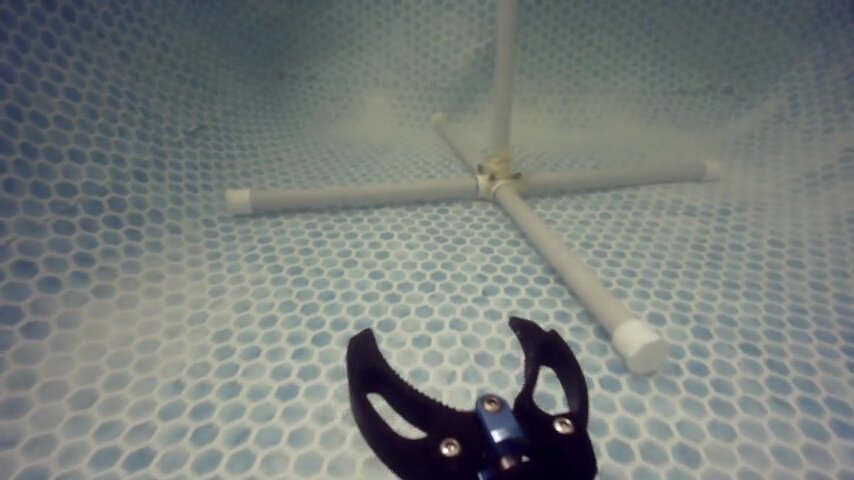}}}$ &
  $\vcenter{\hbox{\includegraphics[width=0.19\textwidth]{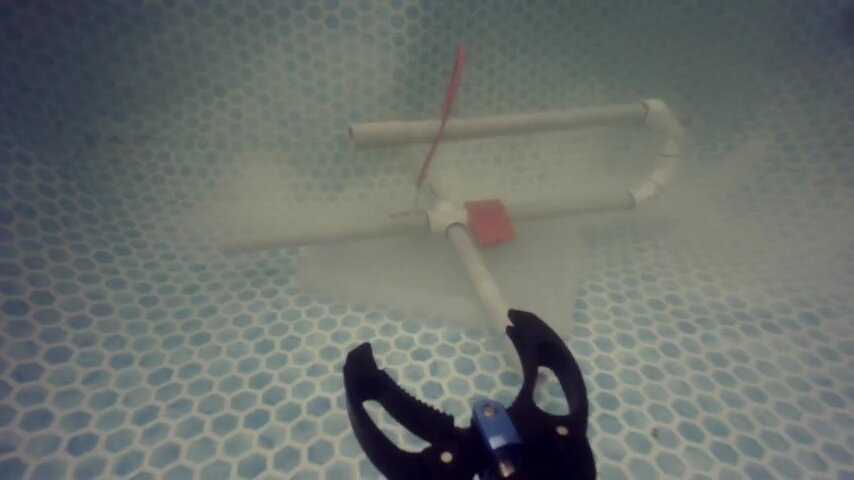}}}$ \\
  \noalign{\vskip 6pt}

  $\vcenter{\hbox{\rotatebox[origin=c]{90}{\textbf{Goal}}}}$ &
  $\vcenter{\hbox{\includegraphics[width=0.19\textwidth]{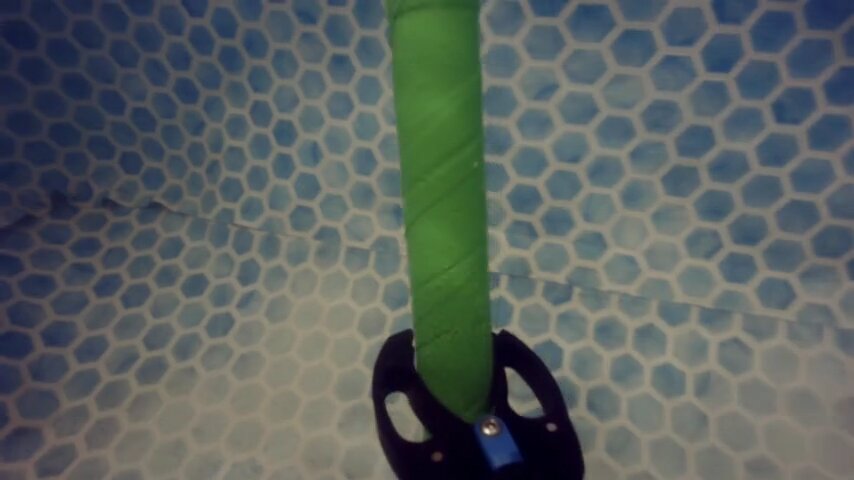}}}$ &
  $\vcenter{\hbox{\includegraphics[width=0.19\textwidth]{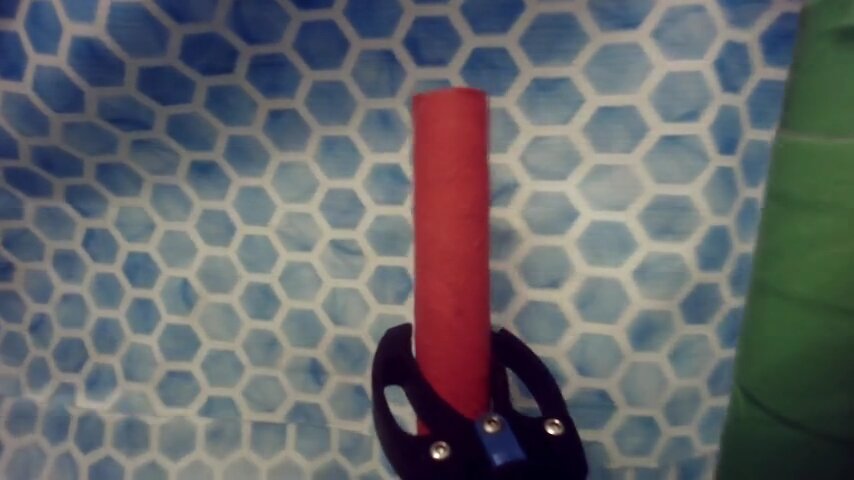}}}$ &
  $\vcenter{\hbox{\includegraphics[width=0.19\textwidth]{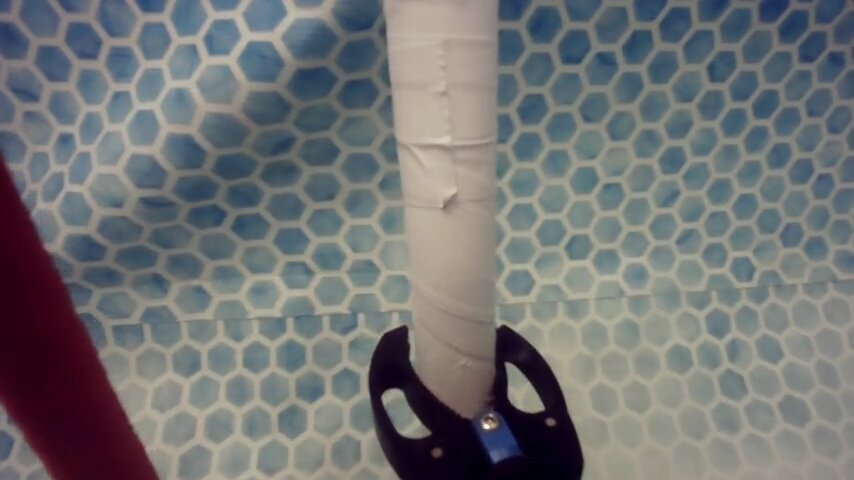}}}$ &
  $\vcenter{\hbox{\includegraphics[width=0.19\textwidth]{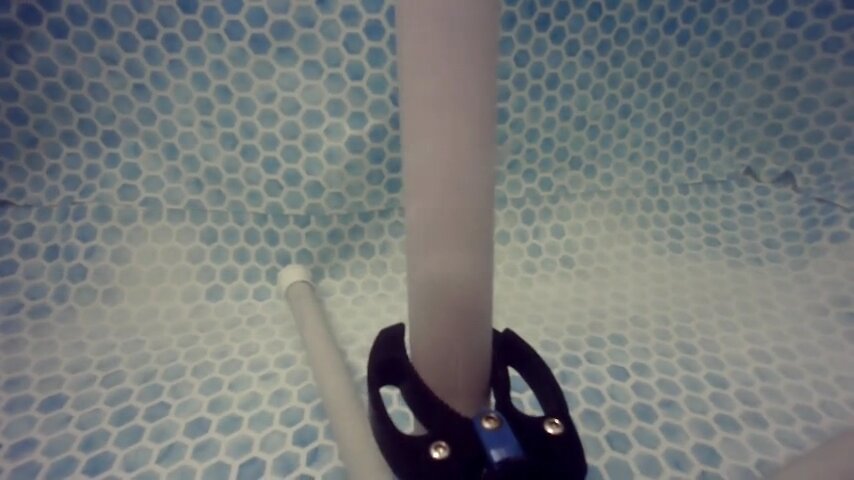}}}$ &
  $\vcenter{\hbox{\includegraphics[width=0.19\textwidth]{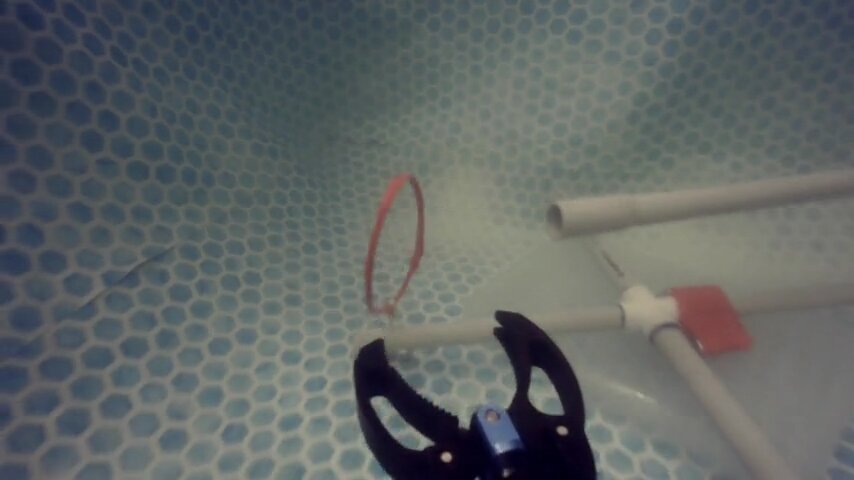}}}$ \\

  \end{tabular}%
  }
  \caption{\textbf{Pool Deployment: Experimental tasks evaluated in a controlled environment.} Data collection in the Pool (Control Environment) was fully illuminated with a split (see Fig.~\ref{fig:pool_dataset}) between calm and turbulent waters created by a human volunteer (see Fig.~\ref{fig:calm_vs_turb}). Rod locations and color order were randomized between episodes. Post location stayed relatively the same, but initial robot and ring configurations varied between episodes as well.}
  \label{fig:pool_tasks}
\end{figure*}

\section{Introduction}

Marine conservationists and archaeologists routinely perform tasks such as environmental monitoring and specimen retrieval, constrained by poor visibility, dynamic currents, and the physiological limits of human divers. At deeper depths, monitoring requires repeated dives that are slow, costly, and risky due to lengthy decompression procedures \cite{deepdives}. Automating these operations, e.g. via robots, could accelerate scientific research and reduce diver risk.

Recent advances in Behavior Cloning (BC) \cite{act} have enabled robust single-task robot policies from relatively little data. While prior works explore the performance of these approaches in aquatic robotics \cite{aquabot,umiunderwater}, deployment in real marine archaeology and conservation missions remains largely unexplored.

BC-based policies, particularly Vision-Language-Action (VLA) models, have shown remarkable generalization to novel tasks and objects, making them an appealing candidate for underwater manipulation. However, they are trained almost exclusively on terrestrial datasets featuring stable lighting and fixed-base manipulators, while underwater vehicles operating in marine environments experience unconstrained 6-DOF motion, drag, turbidity, visual distortion, occlusion, and distractors (see Fig.~\ref{fig:method}). Whether these policies retain their generalization capabilities under such conditions and across diverse archaeological sites remains an open question. 

By enabling increased levels of autonomy for underwater robotic manipulation tasks, there will be scientific value not just within the field of robotics, but as well in the fields of marine archaeology, oceanography, geology, biology, and environmental conservation. This leads to a motivating question: how can the recent advances in terrestrial behavior cloning be leveraged for underwater environments? 

In addressing this driving question, the work presented below yields the following contributions:

\begin{enumerate}
     \item To our knowledge, the world's first instance of successfully implementing a multi-task VLA underwater on real hardware, where a VLA is defined as a single Vision-Language-Action model that can execute multiple tasks and be commanded to execute a specific task via language based inputs.
    \item A real-world case study of BC manipulation models deployed on an underwater Remotely Operated Vehicle (ROV) at a Mediterranean shipwreck for archaeology and conservation tasks.
    \item A thorough investigation into how different proprioception configurations and underwater sensor modalities in a model's observation and action spaces affect generalist model behavior underwater, including a new action space that leverages an underwater robot depth sensor and feedback control for stable motion in the vertical direction.
    \item Introducing the \ours as an open source repository that enables users to collect underwater teleoperation data, train single (ACT) and multi-task models (DynamicVLA, SmolVLA), and run inference on Blue Robotics' BlueROV2 while staying compatible with BlueOS.
\end{enumerate}

\begin{figure}[h]
    \centering
    \includegraphics[width=\columnwidth]{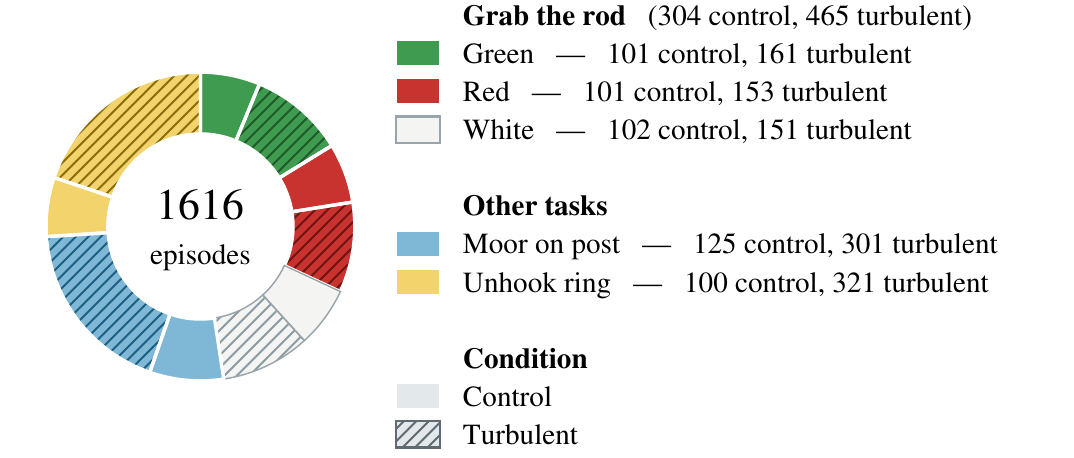}
    \caption{\textbf{Pool Deployment: Dataset split.}
    Training data collected in the pool environment incorporates multi-modal observations and human-made turbulent waters. The dataset is broken down into two conditions: a control condition (calm waters) and a turbulent condition.}
    \label{fig:pool_dataset}
\end{figure}

\begin{figure*}[t!]
\centering
\resizebox{\textwidth}{!}{%
\begin{tabular}{c c c c c c c}

& \textbf{Grab Green Rod} 
& \textbf{Grab Red Rod} 
& \textbf{Grab Red Rod} 
& \textbf{Moor on Ladder} 
& \textbf{Moor on Post} 
& \textbf{Moor on Rope} \\

& \small\textit{(Pool)} 
& \small\textit{(Marina)} 
& \small\textit{(Shipwreck)} 
& \small\textit{(Marina)} 
& \small\textit{(Shipwreck)} 
& \small\textit{(Shipwreck)} \\[0.3em]

$\vcenter{\hbox{\rotatebox[origin=c]{90}{\textbf{Initial State}}}}$ &
$\vcenter{\hbox{\includegraphics[width=0.15\textwidth]{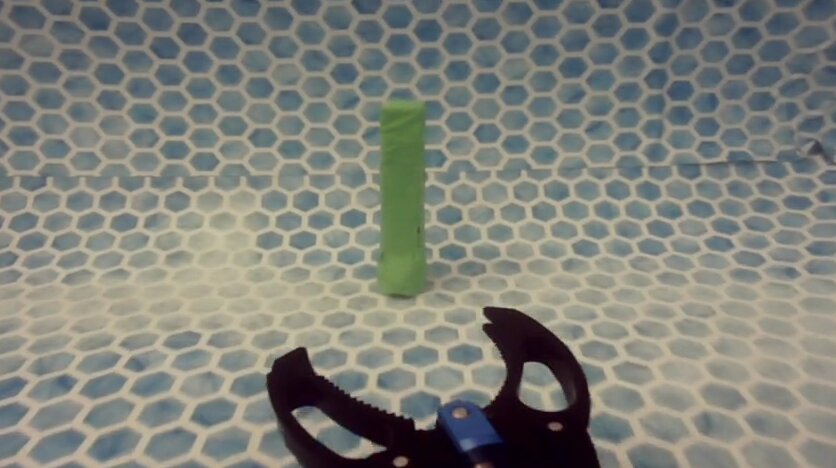}}}$ &
$\vcenter{\hbox{\includegraphics[width=0.15\textwidth]{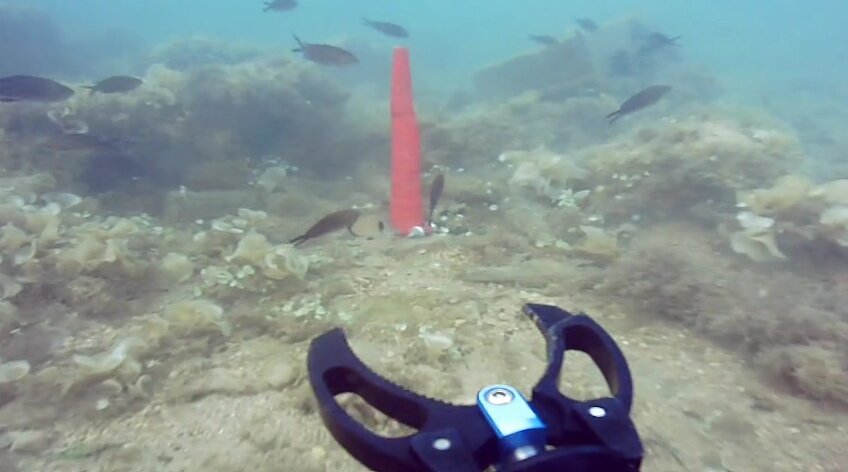}}}$ &
$\vcenter{\hbox{\includegraphics[width=0.15\textwidth]{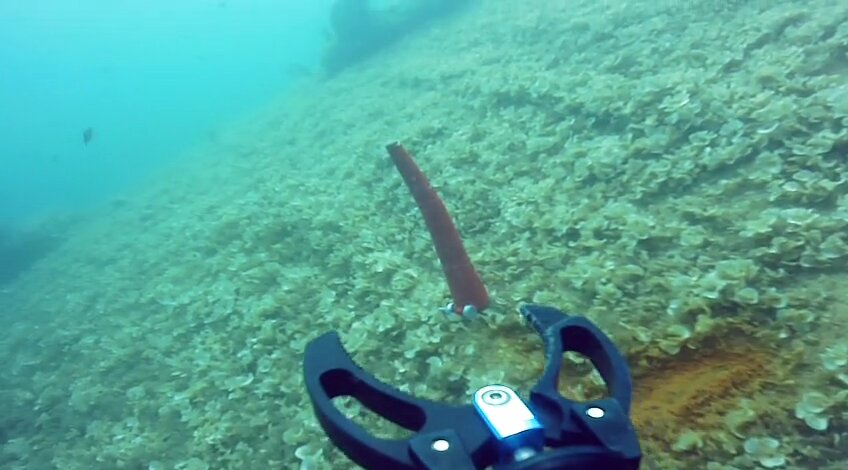}}}$ &
$\vcenter{\hbox{\includegraphics[width=0.15\textwidth]{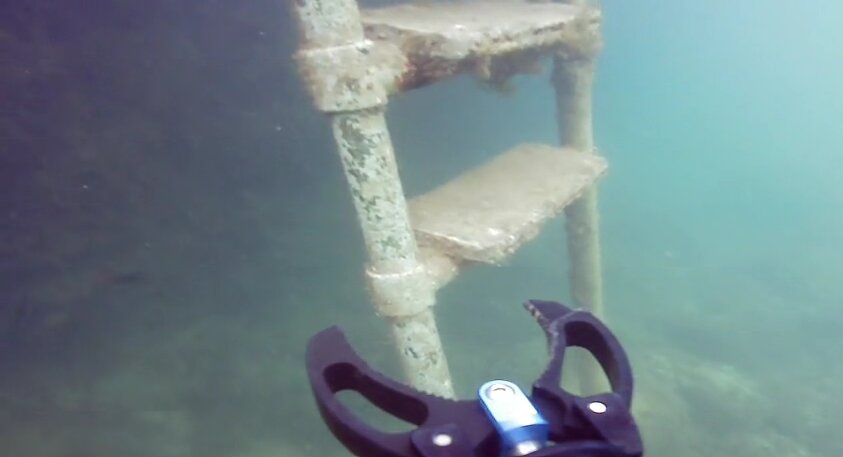}}}$ &
$\vcenter{\hbox{\includegraphics[width=0.15\textwidth]{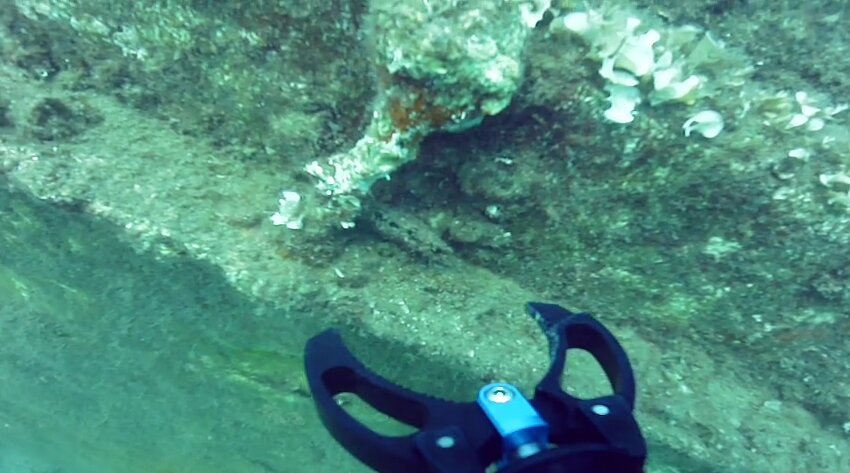}}}$ &
$\vcenter{\hbox{\includegraphics[width=0.15\textwidth]{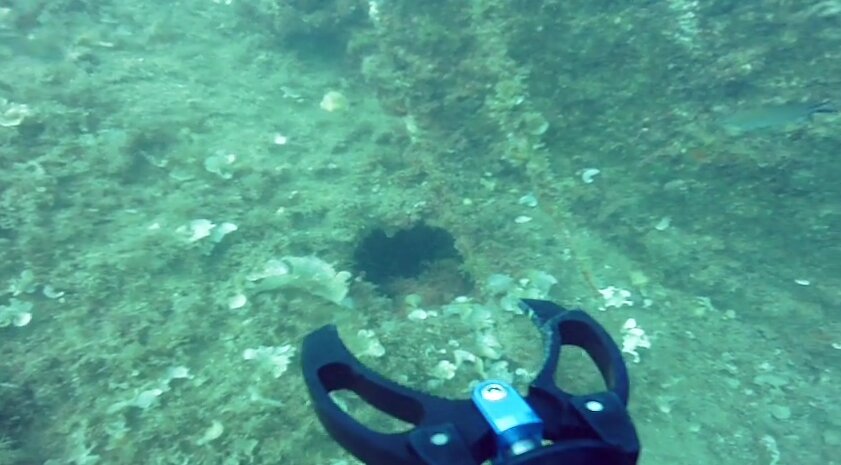}}}$ \\[0.3em]

$\vcenter{\hbox{\rotatebox[origin=c]{90}{\textbf{Goal}}}}$ &
$\vcenter{\hbox{\includegraphics[width=0.15\textwidth]{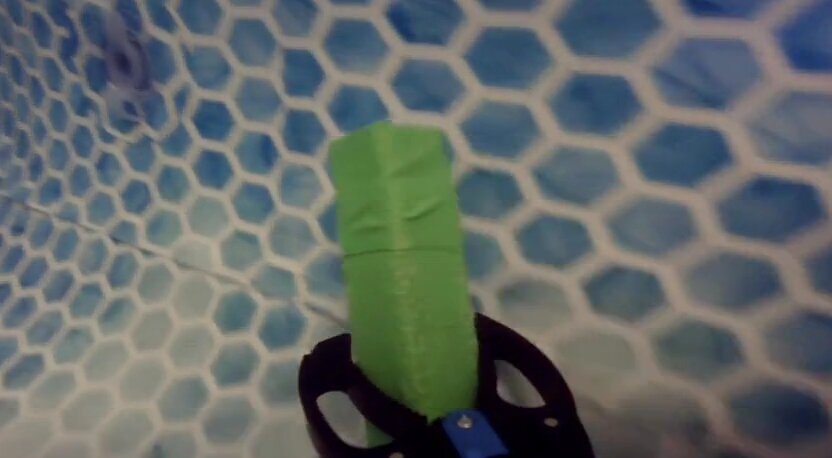}}}$ &
$\vcenter{\hbox{\includegraphics[width=0.15\textwidth]{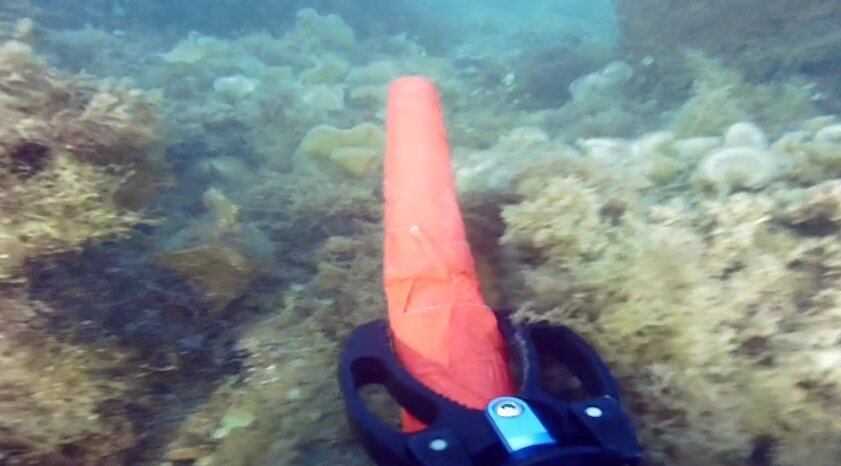}}}$ &
$\vcenter{\hbox{\includegraphics[width=0.15\textwidth]{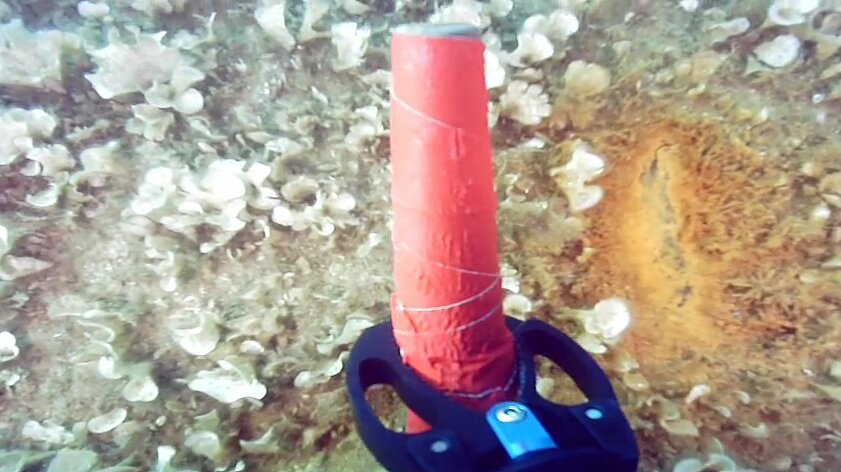}}}$ &
$\vcenter{\hbox{\includegraphics[width=0.15\textwidth]{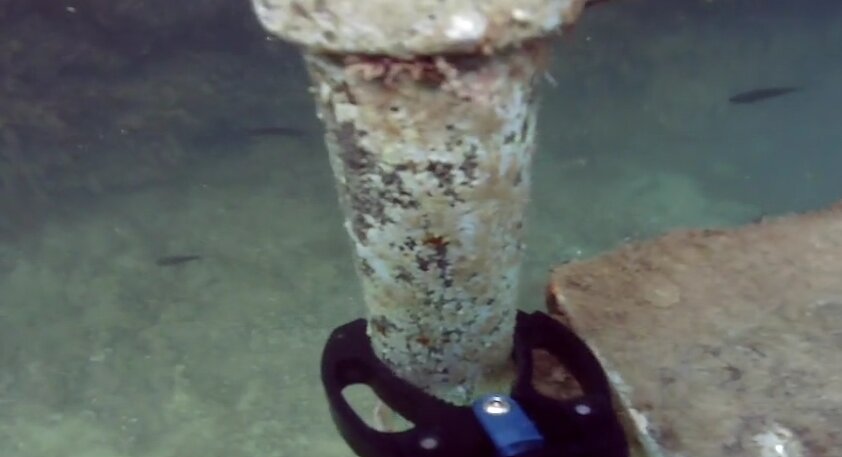}}}$ &
$\vcenter{\hbox{\includegraphics[width=0.15\textwidth]{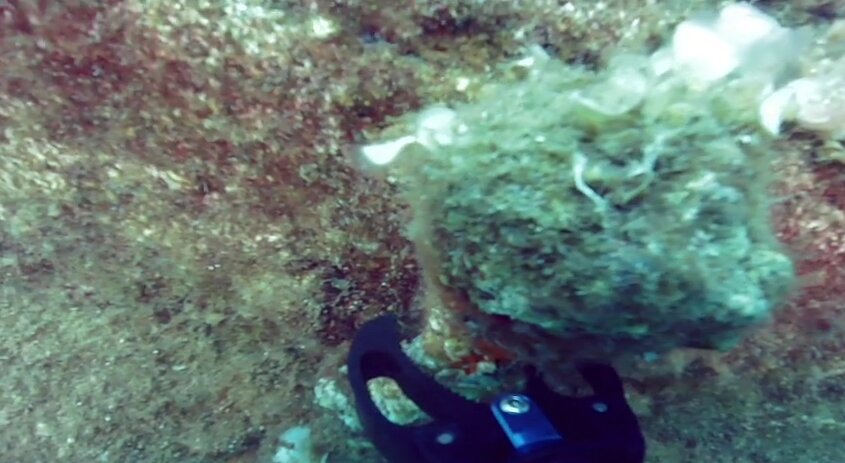}}}$ &
$\vcenter{\hbox{\includegraphics[width=0.15\textwidth]{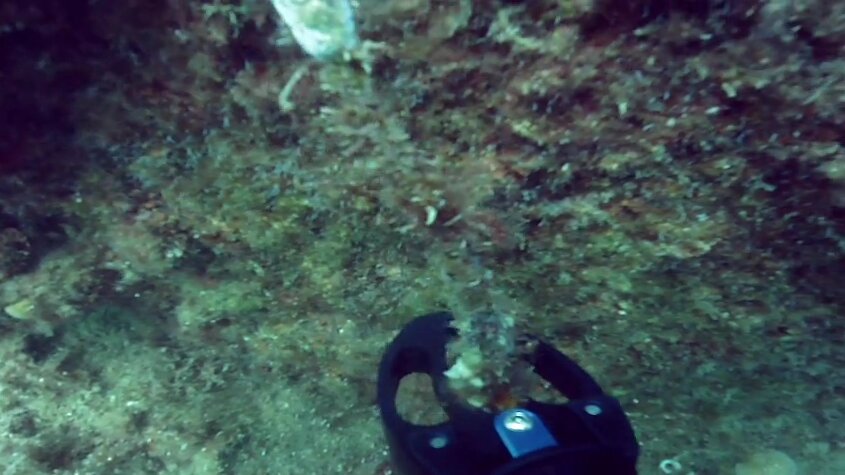}}}$ \\

\end{tabular}%
}

\caption{\textbf{Marine Deployment: Experimental tasks evaluated across different underwater environments.} Data collection in the Pool (Control Environment) was fully illuminated with a split between calm (Control Condition) and turbulent waters. Data collection at the Marina and Coralita X127 shipwreck took place during both day and night, featuring varying disturbances (see Fig.~\ref{fig:method}).}
\label{fig:tasks}
\end{figure*}

\section{Related Work}

\subsection{Autonomy in Underwater Robot Systems}

Underwater robot systems (specifically, ROVs (Remotely Operated Vehicles) and AUVs (Autonomous Underwater Vehicles) \cite{fromrovtoauvs}) have been used for decades, largely for exploration and scientific sampling. However, most AUVs lack manipulation abilities, while many ROVs are equipped with end-effectors. ROV end-effectors are typically remotely operated by humans \cite{oceanone}, while autonomous underwater manipulation has traditionally relied on classical control techniques, including PID, sliding-mode, and model predictive control \cite{autonomousmanipulation}. The machine-learned approaches to developing robot manipulation control policies have largely been utilizing reinforcement learning (RL), e.g. \cite{ROVusingRL, RLunderTorque}, but these works lack real-world ocean deployments.

\subsection{Single-Task BC in Terrestrial Contexts}

BC for single task policies can be described as training a model to output the appropriate signals that can accomplish a single task. The two most notable single task policy models are Action Chunking Transformers (ACT) \cite{act} and Diffusion Policy \cite{diffusionpolicy} (DP). When introduced in \cite{act} the ACT models demonstrated the ability to accomplish highly dexterous tasks with only 100 demonstrations. These models have been augmented in several ways since then, including the addition of mobility \cite{mobileALOHA}. While ACT models have not demonstrated much generalization, DP approaches have often yielded better flexibility but at the cost of needing significantly more demonstrations to achieve high success rates on a task. An advantage of single task policies over most generalist policies is the small size, (e.g. millions of model parameters as opposed to billions of model parameters), which can lead to faster training times and lower inference latency at model deployment. 

\subsection{Generalist Policies in Terrestrial Contexts}

BC for generalist policies can be described as training a single model to be able to accomplish many tasks. Vision Language Action models (VLAs), a broad class of generalist models that take images and text-based language commands as inputs and output robot actions, have become increasingly popular for their robustness. Some of the first VLAs included RT-1 \cite{rt1}, RT-2 \cite{rt2}, and RT-X \cite{rtx}, as well as OpenVLA \cite{openvla}. Later, $\pi_0$ \cite{pizero} from Physical Intelligence demonstrated a multi-expert system that leverages the world generalizability of both Vision Language Models (VLMs) and the action fidelity of diffusion-based policies. A key characteristic of the $\pi_0$ architecture is that it is built upon a large pre-trained VLM backbone (i.e. SigLIP + Gemma), providing cross-attention to another action expert that leverages flow matching to generate robot actions, or controls to be sent to a low-level joint control system. This dual architecture has been mimicked by other research teams such as in \cite{groot} from NVIDIA. Although VLAs have shown strong performance in terrestrial settings, there is no standing literature describing how they perform in a marine environment, and no standing literature successfully implementing VLAs on an ROV.

\subsection{Data Collection Strategies}

The most popular data source for training models continues to be teleoperation, in which a robot manipulator is teleoperated by a human via a leader-follower control system \cite{gello}, or VR system \cite{levr}. These systems generate high-quality data because the data consists of actual robot trajectories in the identical format the models are being trained to generate. Other approaches, like capturing trajectories from videos of humans completing tasks, are typically noisy and require a difficult re-mapping of the human hand’s trajectory to a robot’s end-effector trajectory. However, much progress is being made in this area to increase the efficiency of data collection \cite{egovla}. To note, the data sets generated with large scale human video data have not been collected for use in the underwater domain. When underwater data is collected, it is from human teleoperated demonstrations (e.g., \cite{umiunderwater, aquabot}).

\subsection{Adapting BC to Underwater Contexts}
BC models for robotic manipulation have enabled increasingly generalizable robotic behaviors on terrestrial tasks. Action Chunking Transformers (ACT)\cite{act} learns complex manipulation from low-cost demonstrations, Diffusion Policy \cite{diffusionpolicy}, SmolVLA \cite{smolvla}, and $\pi_0$ \cite{pizero} extend this idea via large-scale pretraining, and DynamicVLA \cite{dynamicvla} improves inference efficiency for dynamic tasks by overlapping inference and execution across action chunks. 

\begin{figure*}[t]
  \centering
  \includegraphics[width=0.98\textwidth]{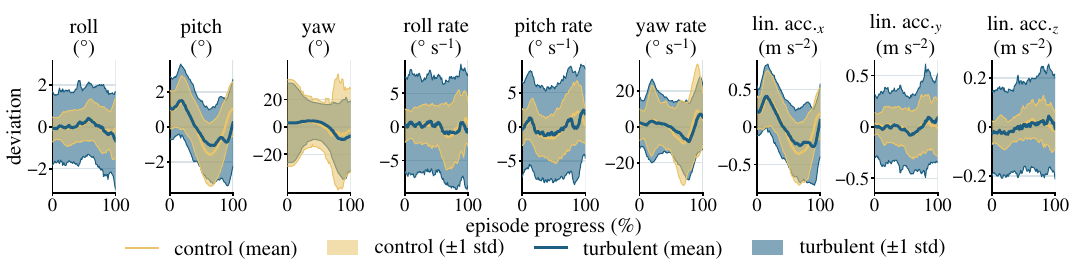}
  \caption{\textbf{Pool Deployment: calm vs.\ turbulent conditions.} Mean attitude and few choice IMU observations plotted over all \textit{Grab the Red Rod} episodes (see Fig.~\ref{fig:pool_dataset}). Since all models (except the baseline) observe roll, pitch, and yaw, these measurements illustrate how turbulence manifests in the observation space.}
  \label{fig:calm_vs_turb}
\end{figure*}

Other works adapt BC beyond their training distribution: AirVLA \cite{pimakeitfly} modifies $\pi_0$ \cite{pizero} actions via a physics-aware, embodiment-specific module, UMI-Underwater \cite{umiunderwater} adapts land-trained diffusion policies underwater with a depth-based affordance predictor and land-data collected from a hand-held device meant to replicate a robot’s end-effector \cite{umi}, and AquaBot \cite{aquabot} shows learning-based policies can improve through underwater data collection. These works demonstrate the utility of BC underwater; however, they all lack underwater demonstrations with full VLA capabilities, such as the ability to command the robot with text based instructions to execute one of many underwater tasks that the single model is capable of. Recent work in \cite{oceansim}, \cite{usim}, and \cite{underwatervla} has attempted to leverage simulation data in the pursuit of underwater BC, given that ocean data is not accessible for many researchers. While synthetic data from simulation also provides another scalable source of underwater training data \cite{oceansim}, many tasks resulting from simulation data suffer from the \textit{sim2real} gap \cite{sim2realVLA}, and typically fails to capture the complexity of the underwater world. Results from \cite{usim}, \cite{underwatervla} are notably not evaluated in the real world. Results from \cite{aquabot} are derived from human teleoperated data and tested in a pool, but are limited to a single task due to the diffusion policy \cite{diffusionpolicy} backbone, and cannot perceive color due to their reliance on depth instead of features. Given the standing literature, fine-tuning multi-task models using underwater data and executing these models underwater to perform complex manipulation tasks remains a critical challenge.

\section{Methodology}
This work presents some of the first applications of BC to underwater environments. Documented below are the different data sets collected and robot control policy model architectures trained with such data, all motivated to provide autonomous control of underwater robots for underwater tasks in underwater environments.

\subsection{Data Collection}
Robot demonstrations were collected across 3 tasks in an in-lab pool, as well as 2 tasks in an ocean setting. 

\subsubsection{Pool Deployment}
To investigate how BC-based policies perform underwater in a controlled environment, three task families were explored (see Fig.~\ref{fig:pool_tasks}): \textit{grab}, approaching an object and grasping it with the end-effector to simulate sampling a specimen, \textit{moor}, approaching a stationary post or rope and latching on with the end-effector to simulate a docking procedure, and \textit{remove}, approaching a ring on a post and unhooking it to simulate a rope line release procedure. 

In total, 1616 human-teleoperated episodes were collected on these tasks (see Fig.~\ref{fig:pool_dataset}), which are split between calm, turbulence free episodes (established as the ''control'') and turbulent episodes created from a human volunteer manually agitating the water to create a sloshing, ''wave'' like motion. 

\subsubsection{Marine Deployment}
To examine the performance of BC-based policies in real ocean conditions, two task families were explored (see Fig.~\ref{fig:tasks}): \textit{grab}, approaching an object and grasping it with the end-effector, and \textit{moor}, approaching a stationary post or rope and latching on with the end-effector to simulate a docking procedure.

In total, 748 human-teleoperated episodes were collected across three sites, progressing from a preliminary controlled test pool set (79 episodes, which is excluded from the Pool Study) to the \textit{Coralita X127} shipwreck (335 episodes total) and finally to Manoel Island Marina, Malta (334 episodes total). To improve dataset robustness, our collection pipeline includes recovery trajectories obtained by resampling directly from failure states. To make our policies more robust for field deployment, both data collection and testing rollouts occurred in the daytime and nighttime (see Fig.~\ref{fig:tasks}).

\begin{figure}[!b]
      \centering
      \includegraphics[width=\columnwidth]{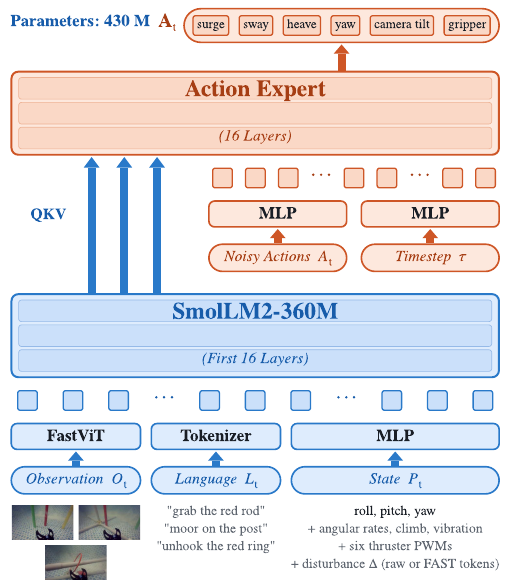}
      \caption{\textbf{Modified DynamicVLA Architecture.} The policy outputs a six-dimensional
      action at each step: four motion commands and two tool commands. The motion
      commands are sent as MAVLink \texttt{MANUAL\_CONTROL} messages to the
      BlueROV2's onboard ArduSub autopilot, which converts them into individual
      thrust commands for its six thrusters. The policy therefore drives the robot
      through the same interface used to collect the teleoperated demonstrations it
      was trained on. Image modified from \cite{dynamicvla}.}
      \label{fig:arch}
  \end{figure}

  \setlength{\tabcolsep}{2pt}

  \begin{table*}[!t]
  \centering
  \caption{\textbf{Pool Deployment: Success rates and forward-pass
  latency of multi-task models.} A success for the \textit{Grab the
  rod} task implies grabbing the right color rod, while a
  \emph{partial} is when the model grabs a rod of the wrong color.
  Including partial successes and across 90 total gradable
  on-hardware rollouts (30 rollouts per task), 10 different DynamicVLA models outperformed
  the baseline. Latency is the time to produce one 20-step action
  chunk (mean over all models: $107.4 \pm 11.4$\,ms). Four models run
  faster than the baseline despite a larger observation space.}
  \label{tab:success}
  \resizebox{\textwidth}{!}{%
  \begin{tabular}{@{}rlr cccc c c cc c@{}}
  \toprule
  & & & \multicolumn{4}{c}{Grab the rod}
    & \begin{tabular}[b]{@{}c@{}}Moor on Post\end{tabular}
    & \begin{tabular}[b]{@{}c@{}}Remove Ring\end{tabular}
    & \multicolumn{2}{c}{Overall}
    & Latency \\
  \cmidrule(lr){4-7}
  \cmidrule(lr){8-8}
  \cmidrule(lr){9-9}
  \cmidrule(lr){10-11}
  \cmidrule(l){12-12}
  \# & Observation Space
    & \begin{tabular}[b]{@{}c@{}}Obs.\\Dim.\end{tabular}
    & succ. & part. & \%
    & \begin{tabular}[b]{@{}c@{}}\%\\w/ part.\end{tabular}
    & \% & \% & \%
    & \begin{tabular}[b]{@{}c@{}}\%\\w/ part.\end{tabular}
    & (ms) \\
  \midrule
  \#1 & none (images only)\textsuperscript{$\dagger$} & 0
    & 8 & 12 & 26.7 & 66.7 & 60.0 & 53.3
    & 46.7 & 60.0 & 101.7 $\pm$ 1.5 \\
  \#2 & attitude, (control dataset) & 3
    & 4 & 8 & 13.3 & 40.0 & 60.0 & 93.3
    & 55.6 & \textbf{64.4} & \textbf{\boldmath 101.0 $\pm$ 0.6} \\
  \rowcolor{black!10}
  \#3 & attitude & 3
    & 10 & 19 & 33.3 & 96.7 & 86.7 & 96.7
    & 72.2 & \textbf{93.3} & \textbf{\boldmath 95.2 $\pm$ 5.0} \\
  \#4 & attitude + $\Delta$ tokens (1.2\,s) & 3
    & 8 & 14 & 26.7 & 73.3 & 63.3 & 46.7
    & 45.6 & \textbf{61.1} & 102.4 $\pm$ 1.2 \\
  \#5 & attitude + $\Delta$ tokens (2.0\,s) & 3
    & 6 & 8 & 20.0 & 46.7 & 50.0 & 93.3
    & 54.4 & \textbf{63.3} & \textbf{\boldmath 100.0 $\pm$ 0.5} \\
  \#6 & attitude + $\Delta$ tokens (4.0\,s) & 3
    & 11 & 16 & 36.7 & 90.0 & 60.0 & 66.7
    & 54.4 & \textbf{72.2} & 102.8 $\pm$ 1.9 \\
  \#7 & attitude + $\Delta$ tokens (6.0\,s) & 3
    & 5 & 18 & 16.7 & 76.7 & 40.0 & 43.3
    & 33.3 & 53.3 & 128.7 $\pm$ 2.8 \\
  \#8 & attitude + raw $\Delta$ & 6
    & 9 & 15 & 30.0 & 80.0 & 36.7 & 76.7
    & 47.8 & \textbf{64.4} & 115.4 $\pm$ 11.6 \\
  \#9 & attitude + IMU & 13
    & 10 & 12 & 33.3 & 73.3 & 73.3 & 66.7
    & 57.8 & \textbf{71.1} & 111.9 $\pm$ 9.9 \\
  \#10 & attitude + IMU + $\Delta$ tokens & 13
    & 12 & 9 & 40.0 & 70.0 & 40.0 & 73.3
    & 51.1 & \textbf{61.1} & 116.4 $\pm$ 12.0 \\
  \#11 & attitude + IMU + raw $\Delta$ & 16
    & 9 & 13 & 30.0 & 73.3 & 30.0 & 76.7
    & 45.6 & 60.0 & 113.5 $\pm$ 11.5 \\
  \#12 & attitude + IMU + thruster PWM & 19
    & 11 & 9 & 36.7 & 66.7 & 46.7 & 66.7
    & 50.0 & 60.0 & \textbf{\boldmath 98.7 $\pm$ 1.6} \\
  \#13 & attitude + IMU + PWM + $\Delta$ tokens & 19
    & 9 & 16 & 30.0 & 83.3 & 50.0 & 93.3
    & 57.8 & \textbf{75.6} & 102.6 $\pm$ 0.7 \\
  \#14 & attitude + IMU + PWM + raw $\Delta$ & 22
    & 10 & 15 & 33.3 & 83.3 & 50.0 & 73.3
    & 52.2 & \textbf{68.9} & 113.9 $\pm$ 11.8 \\
  \bottomrule
  \end{tabular}%
  }
  \par\vspace{2pt}
  \footnotesize\textsuperscript{$\dagger$}Baseline: no proprioception. \textbf{Bold}: better
    than the baseline. Obs. Dims. exclude $\Delta$ tokens.
  \end{table*}

  \setlength{\tabcolsep}{6pt}

\subsection{Model Descriptions}
Unlike fixed-base manipulators, ROVs are continuously jostled by waves and currents across all six degrees of freedom, shifting the camera viewpoint and the vehicle’s pose relative to the target between consecutive actions. As a result, a stationary object to be manipulated in a world frame appears to move in the robot frame. In especially turbulent waters, free-floating manipulation objects tend to sway or move as well, adding another dimension of uncertainty to a policy's predicted trajectory. This turns an otherwise static task into a dynamic one that penalizes any policy too slow to react. As such, three different model architectures were explored, each of which has a promise of faster inference time, a characteristic necessary for dynamic environments. As well, new observation and action spaces are needed for underwater robots. Notably, the use of these observations and actions are novel within the VLA research space.

\subsubsection{Model Architectures}
The three model architectures explored for this work include ACT \cite{act}, SmolVLA \cite{smolvla}, and DynamicVLA \cite{dynamicvla}. Chronologically, ocean deployments occurred before the pool deployments, and all models trained for the ocean were trained on site to reflect realistic field situations. Specifically, one ACT and one SmolVLA policy per task family (from the data set collected) were trained and evaluated on-hardware in the ocean. ACT and SmolVLA were chosen for their ease of use, with ACT being faster to train on site versus SmolVLA.

For in-lab pool experiments, where a more thorough exploration and ablation studies could be performed, the model architecture DynamicVLA \cite{dynamicvla} was chosen due to its low computational overhead and reactivity to target object motion and environmental disturbances. Fig.~\ref{fig:arch} illustrates the DynamicVLA architecture, but with a different observation space and action space.

\subsubsection{Observation Spaces}
While most aspects of the model architectures do not differ from those published in \cite{act,smolvla,dynamicvla}, the observation spaces and even video data content differ greatly. 

\begin{figure*}[!t]
\centering
\includegraphics[width=0.98\textwidth]{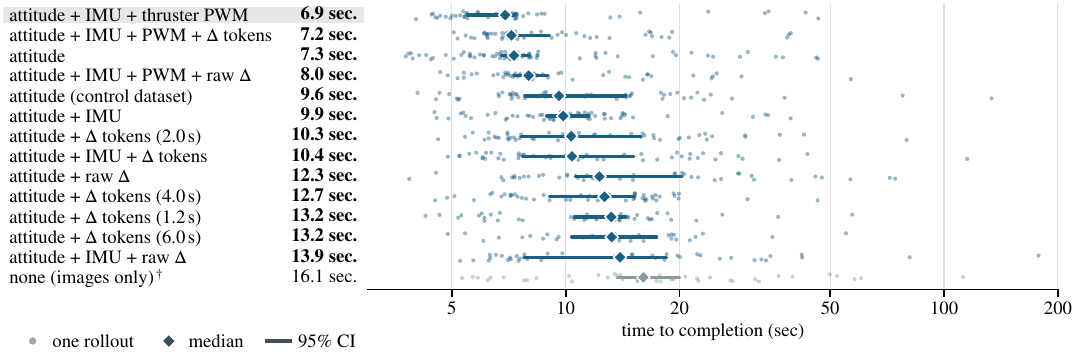}
\caption{\textbf{Pool Deployment: Multi-task models ranked.} The median time to completion was calculated for each model over the successful rollouts, (a subste of the 90 total). All models outperformed the baseline in speediness of a successful task completion. \textsuperscript{$\dagger$}Baseline: no proprioception of any kind.}
\label{fig:rank_speed}
\end{figure*}

For the two model architectures trained for ocean deployment, a six-dimensional observational space was used that includes: roll, pitch, yaw, depth, depth set-point, and the depth controller's output, plus RGB video (see Fig.~\ref{fig:method}). 

For the DynamicVLA models trained for in-lab pool deployments, a greater observation space was considered. At each control step during an episode of data, the ROV reports a 19-channel telemetry row over MAVLink at 50Hz, from which the maximal observation space is $o_t = [\phi, \theta, \psi, a_x, a_y, a_z, \omega_x, \omega_y, \omega_z, v_z, \text{vib}_x, \text{vib}_y, \text{vib}_z, u_1, u_2, u_3,\newline u_4, u_5, u_6]^T$, where:

\begin{enumerate}
    
    \item The ATTITUDE block consists of roll ($\phi$), pitch ($\theta$), and yaw ($\psi$) (rad), forming a 3-dimensional observation space.
    
    \item The IMU block adds ten channels: body-frame linear accelerations $a_x$, $a_y$, and $a_z$ ($\text{m/s}^2$) from raw IMU readings; EKF-fused angular rates $\omega_x$ (\textit{rollspeed}), $\omega_y$ (\textit{pitchspeed}), and $\omega_z$ (\textit{yawspeed}) (rad/s) from ATTITUDE; vertical velocity $v_z$ (\textit{climb}) (m/s) from VFR HUD; and the three vibration measurements $\text{vib}_x$, $\text{vib}_y$, and $\text{vib}_z$ ($\text{m/s}^2$) from VIBRATION. Combined with the ATTITUDE block, this forms a 13-dimensional observation space.

    \item The THRUSTER PWM block adds six commanded thruster outputs $u_1, u_2, u_3, u_4, u_5, u_6$ from the raw server output ($\mu\text{s}$), with a neutral value of 1500~$\mu\text{s}$. Combined with the ATTITUDE and IMU blocks, this forms a 19-dimensional observation space.
\end{enumerate}

From these observations different combinations are selected to evaluate their effect on policy performance.

To investigate how turbulence affects VLA performance, a 3-dim disturbance channel, raw $\Delta_{x,y,z}$, is derived alongside the 19-dim observation space. At each timestep, the ROV's expected acceleration was estimated from its orientation and thruster commands. The difference between this expected acceleration and the measured acceleration was treated as the environmental disturbance:
\begin{equation}
a(t) = G\,\mathrm{orient}(t) + \Phi\,u(t) + \Delta(t),
\end{equation}

\begin{itemize}
    \item $a(t) = [a_x, a_y, a_z]$: measured linear acceleration.
    \item $G\,\mathrm{orient}(t)$: acceleration caused by gravity given the ROV's orientation.
    \item $u(t) = \mathrm{PWM}(t) - 1500$: centered thruster commands.
    \item $\Phi$: mapping from thruster commands to acceleration.
    \item $\Delta(t)$: estimated environmental disturbance.
\end{itemize}

However, a single frame of $\Delta$ carries no turbulence history. Because ocean disturbances are low-frequency and periodic, turbulence only becomes legible over a window of frames. Rather than feed this window in as raw floats, the window is then tokenized with the FAST tokenizer \cite{pizerofast}. This gives the policy a compressed representation of the various frequency components of the perturbations.

14 total DynamicVLA models (see Fig.~\ref{fig:arch}) were trained on the pool data with a variety of observation space combinations (see Table ~\ref{tab:success}), with the smallest observation space being zero (the baseline model) and the largest space being 22 observation space dims. + 39 dims. for $\Delta$ tokens. $\Delta$ token windows were also varied, from 1.2 seconds to 6.0 seconds. The goal of this ablation study is to investigate what telemetry observations best train DynamicVLA (a VLA built solely on terrestrial data) to succeed underwater.

\subsubsection{Action Spaces}
The action space for the models is a 6-dimensional action space defined as $a_t = [v_x, v_y, \omega_z, z_{\text{ref}}, c_{\text{tilt}}, e_{\text{ref}}]^T$where:\begin{itemize}\item Horizontal velocity commands along the body frame $x$ and $y$ axes are given by $v_x$ and $v_y$.\item Yaw rate is denoted by $\omega_z$.\item The vertical altitude setpoint is $z_{\text{ref}}$, which is tracked by a closed-loop PID controller for stable depth-following (defined as a \textit{dive} action).\item The camera tilt toggle command is denoted by $c_{\text{tilt}}$.\item The end-effector setpoint is represented by $e_{\text{ref}}$.\end{itemize}The stable controller compares the depth setpoint $z_{\text{ref}}$ with the depth read by the on-board ROV pressure-based depth sensor (a BAR30) to establish \textit{dive}. 

For pool deployments, the \textit{dive} action was rendered dormant as the pool is shallow, and all pool tasks were performed at equivalent depths. For ocean deployments, the ''dive'' action was active as opposed to dormant. Unlike a pool where the water is shallow, \textit{Coralita X127} is anywhere between 8 meters and 25 meters in depth, and both \textit{grab} and \textit{moor} were performed on an incline and at varying depths. 

\section{Experiment Results and Discussion}\label{sec:results}

\subsection{Hardware and Software}

Experiments used a Blue Robotics BlueROV2 (a vector-configured mini-ROV with six T200 thrusters, see Fig.~\ref{fig:method}) with the Newton Gripper end-effector installation, controlled via a Raspberry Pi 4 companion computer with visual feedback from a front-facing 1080p wide-angle camera. The BlueROV2 is configured with the BlueOS standard operating system. The \ours is used to collect a LeRobot \cite{lerobot} dataset, hand-annotate the data, launch inference through a tethered configuration to computer, record rollout data, and hand-label model rollout successes/partials/failures. The BlueROV2 was controlled using an XBOX controller in the field and through a keyboard in the pool. 
 
\begin{figure*}[h]
    \centering
    \begin{tabular}{@{}c@{\hspace{4pt}}c@{}}
      & {\small Color grabbed during rollout} \\[2pt]
      \rotatebox[origin=c]{90}{\small Color asked for in text prompt} &
      \raisebox{-0.5\height}{\includegraphics[width=0.96\textwidth]{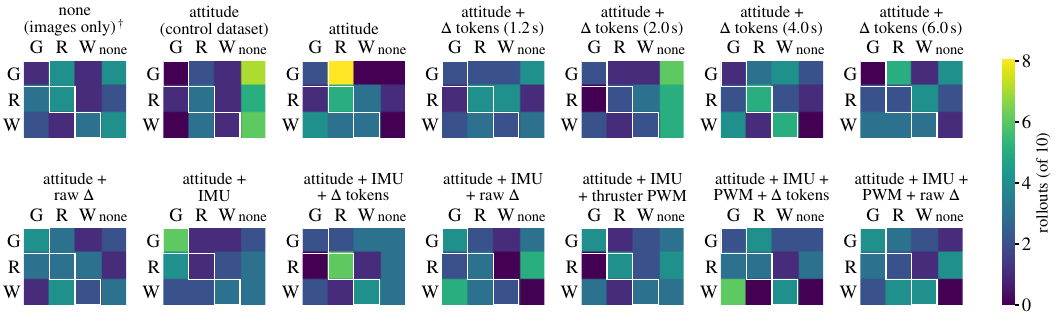}} \\
    \end{tabular}
    \caption{\textbf{Pool Deployment: Rod color confusion per multi-task model.} Perceptive confusion exists strongly in every single multi-task model, and no model stood out to be perceptually more accurate than the other. The most common confusion is between red and green rods. \textit{None} implies that the episode was failed, and hence, no rods were grabbed.
  \textsuperscript{$\dagger$}Baseline: no proprioception of any kind.}
    \label{fig:confusion}
\end{figure*}

\subsection{Pool Deployment Results} 

 Overall, ten models beat the visual-only baseline when partial successes are included, (see Table ~\ref{tab:success}). The highest performer in both overall task success and forward pass latency, i.e. ATTITUDE trained on both control and turbulent pool episodes, achieved a 93.3\% success rate including partial successes, 17.7 percentage points higher than its runner-up, ATTITUDE + IMU + PWM + $\Delta$ Tokens, at 75.6\%. ATTITUDE also has a mean forward pass latency of 95.2 milliseconds, 3.5 milliseconds faster than its runner-up, ATTITUDE + IMU + PWM (see Table ~\ref{tab:success}). ATTITUDE, however, is not the fastest model in terms of successful task time to completion: ATTITUDE + IMU + PWM outperformed ATTITUDE by 0.4 s by a pooled task average.
 
 More proprioception observations did not necessarily lead to better performance: for calm-water manipulation, a smaller observation space provided a better balance between prediction speed and task success, while larger observation spaces, specifically with the addition of IMU and PWM observations, led to faster overall task completion. Similarly, a 4 second $\Delta$ token window enabled better performance than other window lengths for a controlled environment. There is also found no apparent relationship between manipulation success and color confusion, which can imply that differences in manipulation success is not due to the differences in perception between models, but the observations fed into the models directly. This visual confusion in underwater policies is aligned with what was found in \cite{aquabot}, which visual occlusion, warping, and light dissipation can degrade VLA perception underwater. While this can be said about how DynamicVLA performs in our pool in calm waters, limitations of utilizing the pool includes the depth, size, and lack of realistic currents in the pool testing environments.

\subsection{Marine Deployment Results} 

\begin{table}[b]
\centering
\caption{\textbf{Marine Deployment: Success rate by task and location.}
ACT consistently outperforms SmolVLA, and Mooring is the
best-performing task family for both architectures. }
\label{tab:tasklocation}
\footnotesize
{\setlength{\tabcolsep}{3pt}
\begin{tabular}{llrrrr}
\toprule
Task & Location & Policy & $n$ & SR (\%) & 95\% CI \\
\midrule
Grab (red rod)   & Shipwreck & ACT     & 150 & 33.3 & [26.3, 41.2] \\
Grab (red rod)   & Marina    & ACT     & 14  & 42.9 & [21.4, 67.4] \\
Grab (red rod)   & Marina    & SmolVLA & 28  & 0.0  & [0.0, 12.1]  \\
\rowcolor{black!10}
Grab (green rod) & Pool      & ACT     & 22  & 63.6 & [43.0, 80.3] \\
Moor (ladder)    & Marina    & ACT     & 206 & 56.3 & [49.5, 62.9] \\
Moor (ladder)    & Marina    & SmolVLA & 48  & 12.5 & [5.9, 24.7]  \\
\rowcolor{black!10}
Moor (rope)      & Shipwreck & ACT     & 21  & \textbf{66.7} & [45.4, 82.8] \\
Moor (post)      & Shipwreck & ACT     & 19  & 36.8 & [19.1, 59.0] \\
\bottomrule
\end{tabular}}
\end{table}

Our results of the method described in Fig.~\ref{fig:method} can be summarized by Table~\ref{tab:tasklocation}, which reports success by task and location over 508 gradable autonomous policy rollouts.

To test whether a single trained policy could generalize across physically distinct instances of the same task family, one fixed ACT checkpoint was evaluated on three mooring variants without any per-site fine-tuning: on a ladder at the Marina, on a tethered rope on the Shipwreck, and on a post on the Shipwreck (see Fig.~\ref{fig:tasks}). Success rates were 56.3\% (n=206) on the ladder, 66.7\% (n=21) on rope, and 36.8\% (n=19) on the mooring post (Table~\ref{tab:tasklocation}). The comparable success rates on the ladder and rope tasks suggest that the policy learned a site-agnostic mooring strategy rather than overfitting to the Marina ladder's specific visual features. The lower success rate observed for the post likely results from its less forgiving target geometry compared to the ladder and rope, even though both the rope and post exhibited greater occlusion from algae and marine overgrowth than the ladder.

ACT consistently outperformed SmolVLA on every task family (Table~\ref{tab:tasklocation}) pooled across sites: 37.6\% versus 0.0\% on \textit{grab} and 55.7\% versus 12.5\% on \textit{moor}. The gap is largest on the tasks with the most training data: at the Marina ladder, ACT reached 56.3\% on mooring versus SmolVLA's 12.5\%, and 42.9\% versus 0.0\% on \textit{grab red rod}, where SmolVLA never succeeded. ACT's best \textit{grab} result was \textit{grab green rod} in the pool (63.6\%), which is expected given the controlled environment.

When pooled across all tasks and locations, ACT succeeded on 47.9\% of rollouts and demonstrated incredible recovery trajectories during trials, versus 7.9\% for SmolVLA which demonstrated no recovery capabilities. ACT produces short, temporally coherent action chunks that are replanned frequently, keeping execution close to closed-loop. This appears to grant comparative robustness to disturbances that are largely absent from the terrestrial data SmolVLA's backbone was pretrained on. SmolVLA, a substantially heavier model, both replans more slowly and was trained on far less task-specific underwater data than our ACT model, so the performance gap likely reflects inference latency and data quantity rather than a fundamental architectural ceiling. 

The inclusion of ''dive'' in the action space enabled both models to perform tasks at various points on the shipwreck (along the sloping deck and at various points against the hull along the port side), by allowing the ROV to adjust its depth during execution. Sensitivity to depth results in emerging behaviors like recovery maneuvers and greater initial approach precision.

Both tasks evaluated across daylight variations show a consistent performance drop at night. The \textit{moor on ladder} success rate fell from 63.9\% (n=97) in the daytime to 49.5\% (n=109) at night, and the \textit{grab red rod} success rate fell from 42.4\% (n=66) to 26.2\% (n=84) respectively. The drop was larger for grasping, which is consistent with the grasping trajectories being less forgiving to disturbances: unlike mooring, which tolerates vertical error since the ROV can latch anywhere along the ladder or post, grasping the small, swaying rod leaves little margin for error, especially with the reduced contrast and noise of nighttime.



\section{Conclusions and Future Work}\label{sec:future}

This work demonstrates the applicability of VLAs and related models to underwater robot tasks. Results from both pool and marine deployments demonstrate that choices in model architecture and observation space matter when adapting ROVs to VLAs, and VLAs to underwater environments. By experimenting with novel action spaces like depth, models like ACT were able to successfully execute tasks in marine archaeological sites. Observation spaces in multi-task models, such as the addition of $\Delta$ tokens or the simplicity of ATTITUDE, can encourage emerging underwater behaviors like speediness, success rates, or forward pass latency speed. The inclusion of an expanded observation space, therefore, enabled better results than the baseline, despite a backbone trained on terrestrial data alone. This suggests that what we give a policy to observe can be just as important as the model itself, and that an underwater backbone is not necessary for multi-task VLA success. Together, these results suggest that effective underwater VLA deployment depends not only on the choice of model, but on how the model's action and observation spaces are adapted to the physical environment.

The following is recommended for future work: Expanding the \ours repository to include training scripts for a broader set of single and multi-task BC architectures, evaluating said architectures to determine which ones find themselves robust to marine environments, collecting additional trials across \textit{Coralita X127} and other wrecks, evaluating longer-horizon tasks such as tether untangling, and progressively extending deployments to greater depths. In addition, investigating domain generalization (whether data collected on land by terrestrial robotics can be successfully mixed with underwater data for underwater VLA training, and vice versa) is a crucial challenge to confront the difficulty of collecting real world underwater data. An investigation of the application of VLAs in untethered AUVs would also be an interesting challenge which can yield beneficial results for the underwater field robotics community. Together, these directions will move the robotics community toward the broader goal of enabling robust, generalizable, and increasingly autonomous embodied intelligence for real-world underwater robotics.

\section*{Acknowledgment}
The authors would like to thank Heritage Malta for their support during real-world underwater deployments, Dr. Nana Obayashi and the Prema Lab for their generosity in pool access, and Talula Clark for their field assistance.

\balance
\bibliographystyle{IEEEtran}
\bibliography{references}
\balance
\end{document}